\documentclass{article}
\usepackage{spconf}
\usepackage[cmex10]{amsmath}
\usepackage{amsthm}
\usepackage{amssymb}
\usepackage{mathrsfs}
\usepackage{graphicx}
\usepackage{float}
\usepackage{array}
\usepackage{epstopdf}
\usepackage{multirow}

\newcommand{\argmin}{\arg\!\min}

\title{Iris Recognition Using Scattering Transform and Textural Features}
\name{Shervin Minaee, AmirAli Abdolrashidi and Yao Wang}
\address{ECE Department, NYU Polytechnic School of Engineering, USA
\\ \{shervin.minaee, abdolrashidi, yaowang\}@nyu.edu}
\allowdisplaybreaks[1]	
\begin{document}
%
\maketitle
\begin{abstract}
Iris recognition has drawn a lot of attention since the mid-twentieth century. Among all biometric features, iris is known to possess a rich set of features. Different features have been used to perform iris recognition in the past. In this paper, two powerful sets of features are introduced to be used for iris recognition: scattering transform-based features and textural features.
PCA is also applied on the extracted features to reduce the dimensionality of the feature vector while preserving most of the information of its initial value. 
Minimum distance classifier is used to perform template matching for each new test sample.
The proposed scheme is tested on a well-known iris database, and showed promising results with the best accuracy rate of 99.2\%.
\end{abstract}
%
%
\section{Introduction}
\label{sec:intro}
To personalize an experience or make an application more secure and less accessible to  undesired people, we need to be able to distinguish a person from everyone else. It is done using marks from the users to identify them and block unauthorized access, or personalize it based on the trusted identity. To do so, many alternatives are on the table, such as keys, passwords and cards. The most secure options so far, however, are biometric features which cannot be imitated by any other than the desired person himself. They are divided into behavioral features that the person can uniquely create or express, such as signatures, walking rhythm, and physiological characteristics that the person possesses, such as fingerprints and iris pattern. Many works revolve around identification and verification of such data including, but not limited to, fingerprints \cite{Fingerprint}, palmprints \cite{palm_texture}-\cite{palm_wave}, faces \cite{Face} and iris patterns \cite{Iris}.

Iris recognition systems are widely used for security applications, since they contain a rich set of features and do not change significantly over time. They are also virtually impossible to fake.
One of the first modern algorithms for iris recognition was developed by John Daugman and used 2D Gabor wavelet transform \cite{Daugman}. In a more recent work, Kumar \cite{database} proposed to use a combination of Log-Gabor, Haar wavelet, DCT and FFT based features to achieve high accuracy. In \cite{Farouk}, Farouk proposed an scheme which uses elastic graph matching and Gabor wavelet. Each iris is represented as a labeled graph and a similarity function is defined to compare the two graphs. In \cite{Belcher}, Belcher used region-based SIFT descriptor for iris recognition and achieved a relatively good performance. Pillai \cite{Pillai} proposed a unified framework based on random projections and sparse representations to achieve robust and accurate iris matching. The reader is referred to \cite{Iris} for a comprehensive survey of iris recognition.

In most of iris recognition works, the iris region is first detected and the iris is mapped to a rectangular region in polar coordinate. Various iris segmentation algorithms are developed during the past few years \cite{iris_seg}. Foreground segmentation approaches can also be used for iris segmentation \cite{my_seg1}, \cite{my_seg2}. 
It is worth mentioning that no segmentation is performed to extract iris region from the eye image in our work, which makes it much easier to implement.
In this work, two sets of features are extracted from iris images, one of them being the recently introduced set of scattering-transform features and the other one being that of textural features to capture the texture information of irises. We believe that if these features are combined, they will provide a high discriminating power to conduct the recognition task. After the features are extracted, their dimensionality is reduced by applying PCA and then minimum distance classifier is used to recognize new iris images. Skipping the segmentation step makes our algorithm very fast and it can be easily implemented in electronic devices for real time applications using energy-efficient implementation and power management \cite{hosseini}.
This algorithm is tested on the well-known IIT Delhi iris database, and a very high accuracy rate is achieved.  
Three sample iris images of the dataset used in this work are shown in Figure 1.
\begin{figure}[2 h]
\begin{center}
    \includegraphics [scale=0.25] {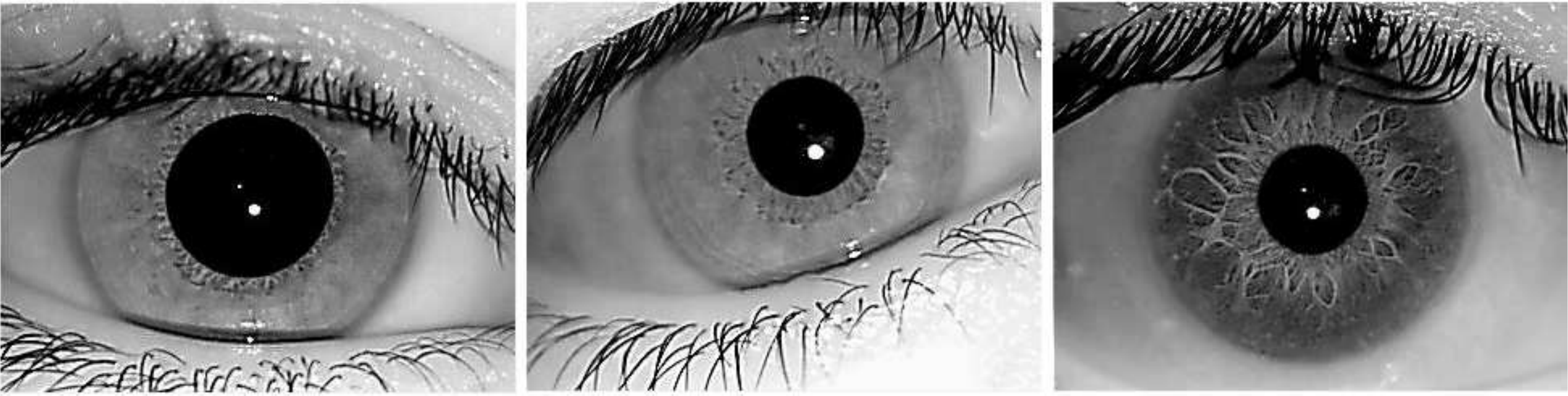}
\end{center}
\vspace{-0.2cm}
  \caption{Three different iris images}
\end{figure}

The rest of the paper is organized as follows. Section \ref{SectionII} describes the features which are used in this work. Section \ref{SectionIII} contains the explanation of the classification scheme. The results of our experiments and comparisons with other works are in Section \ref{SectionIV} and the paper is concluded in Section \ref{SectionV}.

\section{Features}
\label{SectionII}
Extracting good features and image descriptors is one of the most important steps in many computer vision and object recognition algorithms. As a result, many researchers have focused on designing useful features which can be used for a variety of object recognition and image classification tasks. A good feature should have some degree of invariance with respect to translation, slight rotation and deformation. There are many popular features and image descriptors which are being used today, including scale invariant feature transform (SIFT), histogram of oriented gradient (HOG), bag of words (BoW) \cite{sift}-\cite{BoW}, etc.
Geometrical features and sparsity-based features are also used for some biometric and medical applications in several works \cite{chromosome}-\cite{hojjat2}. A new algorithm for feature selection for small datasets is presented in \cite{mtbi}.
Recently, unsupervised feature learning algorithms have been in the spotlight, where the image is fed directly as the input to the deep neural network and the algorithm itself finds the best set of features from the image.

For iris recognition, various features have been used by several researchers, including wavelet-based features, PCA and LDA. 
In this paper, a combined set of two features is used: some derived from the scattering transform, and the rest from the textural information of iris patterns. These features are introduced in the following subsections in more detail.

\subsection{Scattering Features}
The scattering operator is a locally translation-invariant descriptor which is proposed by Stephane Mallat and has achieved state-of-the-art recognition accuracy in several computer vision \cite{mallat1} and audio classification \cite{mallat2} problems. A scattering transform computes local image descriptors with a cascade of three operations: wavelet decompositions, complex modulus and a local averaging. The scattering coefficients are similar to those of the SIFT descriptor, but they contain more high-frequency information than SIFT.
As discussed in \cite{mallat1}, other image descriptors such as SIFT and multiscale Gabor textons can be obtained by averaging the amplitude of wavelet coefficients, calculated with directional wavelets. This averaging provide some sort of local translation invariance, but it also reduces the high-frequency information. Scattering transform recovers part of the high-frequency information lost by this averaging with co-occurrence coefficients having the similar invariance as those of the scattering transform.

In most object recognition tasks, locally invariant features are preferred, since they provide robust representation of images. They can be seen as the averaged value of gradient orientation. Using this averaging, some local deformation and translation will be tolerated. However, such process will reduce too much high-frequency information and therefore could greatly decrease discriminating capability. The scattering features provide richer descriptors for complex structures such as corners and multiscale texture variations.

The scattering operator is designed in a way that it preserves the locally invariance property of SIFT, but it also recovers the lost high-frequency content of the images. Suppose we have a signal $f(x)$. The first scattering coefficient is the average of the signal and can be obtained by convolving the signal with an averaging filter $\phi_J$ as $f*\phi_J$. The scattering coefficients of the first layer can be obtained by applying wavelet transforms at different scales and orientations, and taking the magnitude and convolving it with a low-pass filter $\phi_J$ as shown below:
\begin{gather}
|f* \psi_{j_1,\lambda_1}|*\phi_J
\end{gather}
where $j_1$ and $\lambda_1$ denote different scales and orientations and $j_1<J$. Note that removing the complex phase of wavelet will make these coefficients insensitive to local translation.

Now to recover the high-frequency information, which is eliminated from the wavelet coefficients of first layer by averaging, we can convolve $|f* \psi_{j_1,\lambda_1}|$ by another set of wavelet at scale $j_2<J$, taking the absolute value of wavelet and taking the average:
\begin{gather}
||f* \psi_{j_1,\lambda_1}|*\psi_{j_2,\lambda_2}|*\phi_J
\end{gather}
One can show that $|f* \psi_{j_1,\lambda_1}|*\psi_{j_2,\lambda_2}$ is negligible at scales where $2^{j_1} \leq  2^{j_2}$. Therefore the coefficients are calculated only for $j_1 >j_2 $.

The convolution with $\phi_J$ at the second layer removes high frequencies and yields second-order coefficients which are locally invariant to translation. These high-frequency information can be restored again by finer scale wavelet coefficients in the next layers. 
To obtain the scattering coefficients at the $k$-{th} layer, we have to perform the following procedure iteratively $k$ times:
\begin{gather}
\underset{\ \ \ \ \ \ \ \ \ \ \ \ \ \ \ \ \ \ \ \ \ \ \ \ \ \ j_k<...<j_2<j_1<J, \ (\lambda_1,...,\lambda_k) \in \Gamma^k  }{S_{k,J}(f(x)))= ||f* \psi_{j_1,\lambda_1}|*...*\psi_{j_k,\lambda_k}|*\phi_J}
\end{gather}

The output of scattering transform of the $k$-{th} layer has a size of $p^k {J \choose k}$ where p denotes the number of different orientations. In other words there are $p^k {J \choose k}$ transformed images at the output of the $k$-{th} layer.

The transformed images of the first and second layers of scattering transform for a sample iris image are shown in Figures 2 and 3. These images are derived by applying bank of filters of 5 different scales and 6 orientations. Scattering vector can be thought of as the cascade of convolution with wavelets, non-linear modulus and averaging operators which makes it very similar to the deep convolutional neural network \cite{mallat3}.

\begin{figure}[2 h]

\begin{center}
    \includegraphics [scale=0.25] {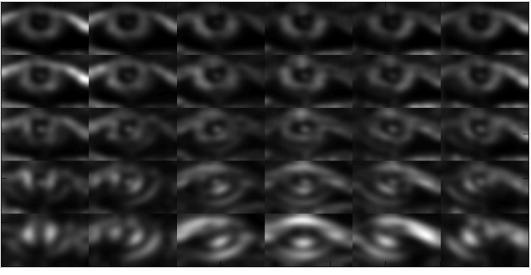}
\end{center}
  \vspace{-0.2cm}
  \caption{The images from the first layer of scattering transform}
\end{figure}
\begin{figure}[2 h]
\begin{center}
    \includegraphics [scale=0.25] {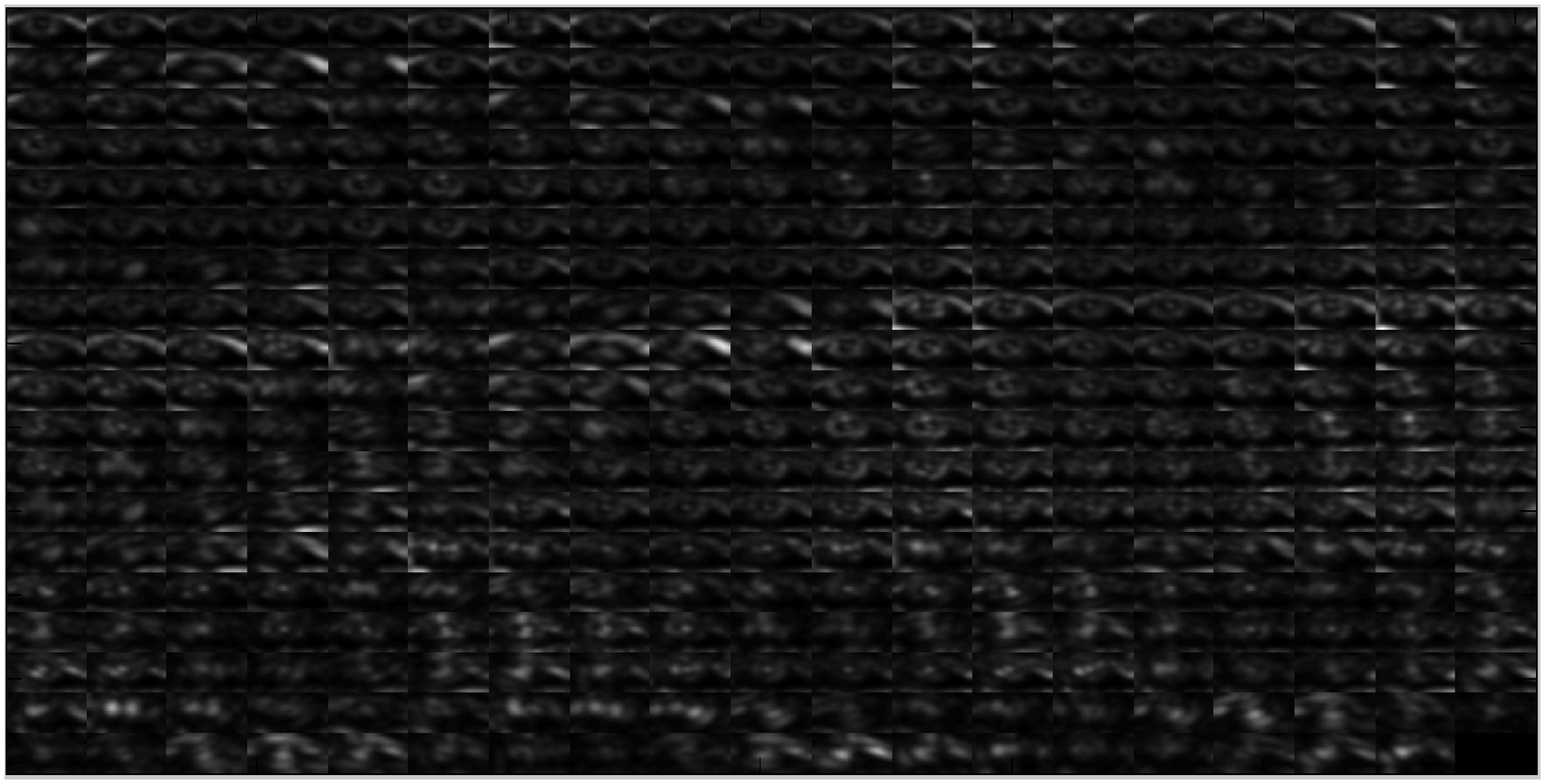}
\end{center}
\vspace{-0.2cm}
  \caption{The images from the second layer of scattering transform}
\end{figure}

To derive scattering features, the scattering-transformed images of all layers up to $m$ are taken and the mean and variance of these images are calculated as scattering features which results in a vector $\textbf{f}_{\textbf{s}}$ of size $\sum_{k=0}^m{2p^k {J \choose k}}$.
For mainstream applications, using two or three levels of scattering transform will be enough.

\subsection{Textural Features}
Textural, spectral and contextual features are the three fundamental pattern elements in recognition. In human interpretation of color photographs, textural features contain the spatial information of intensity variation in a single band \cite{haralick}. To mimic the human visual system, there are several features introduced to capture textural information of an image. Among them, Haralick features and local binary pattern (LBP) are two major groups of textural features. 
Haralick textural features are derived from the co-occurrence matrix of image. Local binary pattern are derived based on the relative comparison between each pixel and its neighboring pixels. 
There are also various modified versions of LBP features such as transition local binary patterns, direction-coded local binary patterns, volume local binary pattern (VLBP).
In this work, Haralick features are used to capture textural information of the image. To extract Haralick features, we first need to derive the co-occurrence matrix. The co-occurrence matrix measures the distribution of co-occurring intensity values for a given offset. If we represent the image as a two-dimensional function which maps pairs of coordinates to the intensity values, i.e., $I: X\times Y \rightarrow G$, where $X=\{1,2,3,...,N_x\}$ and $Y=\{1,2,3,...,N_y\}$, and $G$ denotes the set of all possible grayscale levels. 
Then the co-occurrence matrix $P$ of image $I$ with the offset $(\Delta_x,\Delta_y)$ or $P_{\Delta_x,\Delta_y}(i,j)$can be defined as:
\begin{gather}
 \sum_{m=1}^{N_x} \sum_{n=1}^{N_y} \delta(I(m,n)-i)~ \delta(I(m+\Delta_x,n+\Delta_y)-j)
\end{gather}
where $\delta(x)$ denotes the discrete Dirac function. 
It should be noted that the co-occurrence matrix has a size of ${N_g} \times {N_g}$, where $N_g$ denotes the number of gray levels in the image. 
The offset $(\Delta_x,\Delta_y)$ depends on the direction $\theta$ which can be defined as:
\begin{gather}
\theta= tan^{-1}(\frac{\Delta_y}{\Delta_x})
\end{gather}
Here we have derived the co-occurrence matrix for the offset $(\Delta_x,\Delta_y)=(1,0)$.
In our work, the textural features are extracted on block level. Each image is divided into non-overlapping blocks of size $N \times N$ and their co-occurrence matrices are derived and 14 features are extracted from them. More details about the derivation of these 14 textural features from co-occurrence matrix is provided in the appendix.
Then the features from different blocks are concatenated and formed a longer feature vector.
If an image has a size of $s_1 \times s_2$, the total number of textural features will be:
\begin{gather*}
M=\frac{14s_1s_2}{N^2}
\end{gather*}

In our work the textural features are derived in a slightly different way from the original paper \cite{haralick}, but they are very similar.
Here the co-occurrence matrix is found only for a single pixel horizontal shift (corresponding to $\theta=0$).

After derivation of the set of scattering and textural features, we can concatenate them to form the feature vector of each iris image as: $\textbf{f}=[\textbf{f}_{\textbf{s}}^\intercal, \textbf{f}_{\textbf{t}}^\intercal]^\intercal$, where $\textbf{f}_{\textbf{s}}$ and $\textbf{f}_{\textbf{t}}$ denote the scattering and textural features respectively.

\subsection{Principal Component Analysis}
Principal component analysis (PCA), also known as Karhunen-Loeve transformation, is a powerful algorithm used for dimensionality reduction \cite{PCA}. Given a set of correlated variables, PCA transforms them into another domain such the transformed variables are linearly uncorrelated. This set of linearly uncorrelated variables are called principal components. PCA is usually defined in a way that the first principal component has the largest possible variance and the second one has the second largest variance and so on. Therefore after applying PCA, we could only keep a subset of principal components with the largest variance to reduce the dimensionality. PCA can be thought of as fitting a $k$-dimensional ellipsoid to a set of data, where each axis of the ellipsoid represents a principal component. There are also others dimensionality reduction algorithms which are designed based on PCA such as kernel-PCA and sparse-PCA.
PCA has many applications in computer vision. Eigenface is one representative application of PCA in computer vision, where PCA is used for face recognition \cite{eigenface}.

Without going into too much detail, let us assume we have a dataset of $N$ iris images and $\{f_1,f_2,...,f_N\}$ denote their features. Also let us assume that each feature has dimensionality of $d$. To apply PCA, all features need to be centered first by removing their mean:  $z_i= f_i- \bar{f}$ where $\bar{f}= \frac{1}{N} \sum_{i=1}^N f_i$.
Then the covariance matrix of the centered images is calculated:
\begin{gather}
C= \sum_{i=1}^N z_i z_i^T 
\end{gather}
Next the eigenvalues $\lambda_k$ and eigenvectors $\nu_k$ of the covariance matrix $C$ are computed. Suppose $\lambda_k$'s are ordered based on their values. Then each $z_i$ can be written as $z_i= \sum_{i=1}^d \alpha_i \nu_i$.
We can reduce the dimensionality of the data by projecting them on the first $K (\ll d)$ principal vectors as:
\begin{gather*}
\hat{z_i}= (\hat{z_1},\hat{z_2},...,\hat{z_K})= (\nu_1^{T} z_i, \nu_2^{T} z_i,..., \nu_K^{T} z_i)= (\alpha_1,...,\alpha_K)
\end{gather*}
By keeping $k$ principal components, the percentage of retained variance can be found as: $\frac{\sum_{i=1}^k \lambda_i}{\sum_{i=1}^d \lambda_i }$. Hence one simple way to choose $k$ would be to pick a value such that the above ratio is less than $\epsilon$, where $\epsilon$ is usually chosen between 95\% to 99\%.

\section{Recognition Algorithm: Minimum Distance Classifier}
\label{SectionIII}
There are various classifiers which can be used for this task, including majority voting algorithm, support vector machine, neural network and minimum distance classifier. In this work minimum distance classifier has been used 
which is quite popular for template matching problems. One benefit of minimum distance classifier is that it does not need any training, making it much faster than most of the other classifiers. As long as the features are discriminative enough to separate different classes, the minimum distance classifier will provide high accuracy, otherwise using other classifiers would be a better option. Minimum distance classifier finds the distance between the features of the training samples and those of an unknown subject, and picks the training sample with the minimum distance to the unknown as the answer. To put it in equation, if we show the features of the test subject as $F^*$ and those of the test sample $i$ with $F^{(i)}$, the test subject is matched to the sample that satisfies the following:
\begin{gather}
i^*=\argmin_i \big[ dis(F^*,  F^{(i)}) \big]
\end{gather}
We have used Euclidean distance as our distance metric.

\section{Experimental results and analysis}
\label{SectionIV}
This section presents a detailed description of experimental results.
Before showing the results, let us describe the parameter values of our algorithm.
For each image, scattering transform is applied up to two levels with a set of filter banks with 5 scales and 6 orientations, resulting in 391 transformed images. The mean and variance of each scatter-transformed image are used as features, resulting in 782 scattering features. The scattering features are derived using the software implementation provided by Mallat's group \cite{scat}.

To extract textural features, each image is divided into 12 smaller blocks and 14 features are derived from any of them which results in a total of 168 textural features. Therefore the concatenated feature vector has a length of 950. 
Then PCA is applied to all features and the first 80 PCA features (retain above 99\% of the initial features' energy) are used for recognition. Minimum distance classifier is used for template matching.

We have tested our algorithm on a popular iris database collected by IIT Delhi. This database contains 2240 iris images captured from 224 different people.
The images of 21 people (around 10\%) are used as a validation set to find the optimum value of the parameters of the algorithm, and the rest are used for evaluation. For each person, around half of the images are used for training and the rest for testing.


Figure 4 shows the recognition accuracy using different numbers of PCA features.
Interestingly, even by using few PCA features, we are able to get a very high accuracy rate. As it can be seen, using 80 PCA features results in a accuracy rate above 99\%, which will not increase by using more PCA features.
\vspace{-0.2cm}
\begin{figure}[3 h]
\begin{center}
    \includegraphics [scale=0.42] {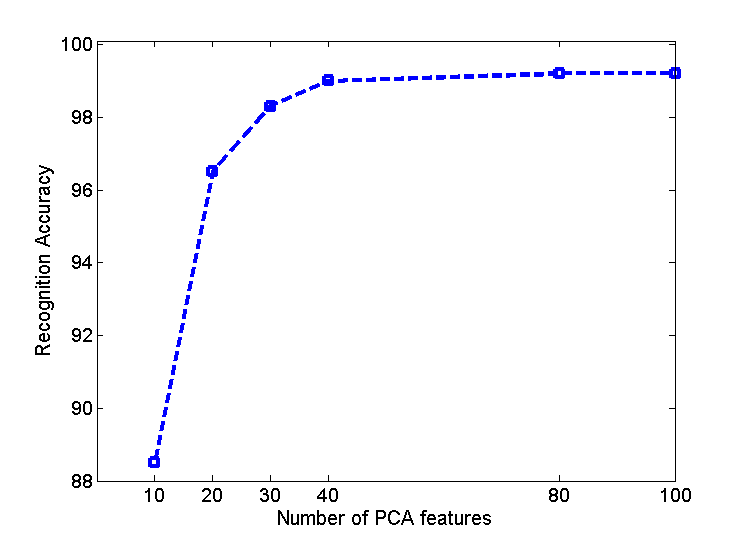}
\end{center}
  \vspace{-0.3cm}
  \caption{Recognition accuracy as a function of number of PCA features}
\end{figure}

Table 1 provides a comparison of the performance of the proposed scheme and those of  other recent algorithms. The accuracy of the proposed scheme is reported as the highest rate achieved by 80 PCA features. As it can be seen, using the combination of scattering and textural features, we are able to outperform previous approaches. This is mainly due to the richness of both scattering and Haralick features which are able to capture high-frequency patterns of irises, providing a very high discriminating power. One main advantage of this scheme is that, it does not require segmentation of iris from eye images (although the segmentation could improve the results for some difficult cases).

\begin{table} [h]
\centering
  \caption{Comparison with other algorithms for iris recognition }
  \centering
\begin{tabular}{|m{5.8cm}|m{2cm}|}
\hline
\ \ \ \ \ \ \ \ \ \ \ \ \ \ \ \ \ \ \ \ \ \ \ \ \ Method &  Recognition rate\\
\hline
Haar Wavelet \cite{database} & \ \ \ \ \  96.6\% \\
\hline
Log Gabor Filter by Kumar \cite{database} & \ \ \ \ \ 97.19\% \\
\hline
Fusion \cite{database} & \ \ \ \ \  97.41\% \\
\hline
Elastic Graph Matching \cite{Farouk} & \ \ \ \ \ 98\% \\
\hline
Proposed scheme using 80 PCA features & \ \ \ \ \  99.2\% \\
\hline
\end{tabular}
\label{TblComp}
\end{table}

The experiments are performed using MATLAB 2012 on a laptop with Core i5 CPU running at 2.6GHz. The execution time for the proposed scheme is about 11ms for each image which is fast enough to be used for real-time applications.

\section{Conclusion}
\label{SectionV}
This paper proposed a set of scattering and textural features for iris recognition. The scattering features are extracted globally, while the textural features are extracted locally. Scattering features are locally invariant and carry a great deal of high-frequency information which are lost in other descriptors such as SIFT and HOG. The high-frequency information provides great discriminating power for iris recognition.
Principal component analysis is applied on features to reduce dimensionality. Then minimum distance classifier is used to match new iris images with training images.
This algorithm is tested on a well-known dataset, and a high accuracy rate is achieved which outperforms the previous best results achieved on this dataset. In the future, we will investigate to apply the proposed set of features to more challenging iris datasets and also other biometric recognition problems.

\section*{Acknowledgments}
The authors would like to thank Stephane Mallat's research group at Ecole Normale Superieure for providing the software implementation of scattering transform. We would also like to thank IIT Delhi for providing the iris database \cite{database}.

\section*{Appendix. More details on Haralick textural features}
To find textural features from the co-occurrence matrix, we first need to find the following terms which are used for derivation of Haralick features. $p(i,j)= P(i,j)/R$ denotes the normalized co-occurrence matrix which can be thought as a probability distribution. $p_x(i)$ and $p_y(j)$ denote the marginal probabilities along $x$ and $y$. $p_{x+y}(k)$ and $p_{x-y}(k)$ denote the probabilities of $x+y$ and $x-y$. $HXY$ denotes the entropy of $p(i,j)$ and:
\begin{align*}
&HXY1= -\sum_{i}\sum_{j} p(i,j) \log\big(p_x(i)p_y(j)\big) \\
&HXY2= -\sum_{i}\sum_{j} p_x(i)p_y(j) \log\big(p_x(i)p_y(j)\big) \\
&Q(i,j)= \sum_{k} \frac{p(i,k)p(j,k)}{p_x(i)p_y(k)}
\end{align*}
Using the above terms, the following 14 textural can be derived for each image:
\begin{align*}
&f_1= \sum_{i}\sum_{j} \big[p(i,j) \big]^2,  \ \ \ \ \ \ \  \ \ \ \ \ \ \ \ \ \ \text{Angular Second Moment} \\
&f_2= \sum_{k=0}^{N_g-1}k^2 p_{x-y}(k), \  \ \ \ \ \ \ \ \ \ \ \ \ \ \ \ \  \ \ \text{Contrast} \\
&f_3= \frac{\sum_{i}\sum_{j}ijp(i,j)-\mu_x\mu_y}{\sigma_x\sigma_y}, \ \ \ \ \ \text{Correlation} \\
&f_4= \sum_{i}\sum_{j} (i-\mu)^2p(i,j), \ \ \ \ \ \ \ \ \ \ \text{Variance} \\
&f_5= \sum_{i}\sum_{j}  \frac{1}{1+(i-j)^2} p(i,j), \ \ \ \text{Inverse Diference Moment}\\
&f_6= \sum_{k=2}^{2N_g}k p_{x+y}(k), \ \ \ \ \ \ \ \ \ \  \ \ \ \ \ \ \ \ \ \ \ \ \ \  \text{Sum Average} \\
&f_7= \sum_{k=2}^{2N_g}(k-f_6)^2 p_{x+y}(k), \  \ \ \ \ \ \ \ \ \ \ \ \text{Sum Variance} \\
&f_8= -\sum_{k=2}^{2N_g} p_{x+y}(k) \log(p_{x+y}(k)), \ \text{Sum Entropy} \\
&f_9= -\sum_{i}\sum_{j} p(i,j) \log(p(i,j)), \  \ \text{Entropy} \ \\
&f_{10}= \sum_{k=0}^{N_g-1}(k-\mu_{x-y})^2 p_{x-y}(k),  \ \ \ \ \text{Diference Variance} \\
&f_{11}= -\sum_{k=0}^{N_g-1} p_{x-y}(k) \log p_{x-y}(k), \ \text{Difference Entropy} \\
&f_{12}= \frac{HXY-HXY1}{max\{HX,HY\}}  \\~\\
&f_{13}= \sqrt{1-exp{[-2(HXY2-HXY)]}}  \\~\\
&f_{14}= \sqrt{\text{Second largest eigenvalue of} \ Q}  \\
\end{align*}

\end{document}